\documentclass[letterpaper, 10 pt, journal, twoside]{IEEEtran}

\usepackage{amsmath,amsfonts}
\usepackage{array}
\usepackage[caption=false,font=normalsize,labelfont=sf,textfont=sf]{subfig}
\usepackage{textcomp}
\usepackage{url}
\usepackage{verbatim}
\usepackage{graphicx}
\usepackage{cite}
\usepackage{newtxtext,newtxmath}
\usepackage{multirow}
\usepackage{tikz}
\usepackage{xcolor}
\usepackage{bm}
\usepackage[linesnumbered,ruled,vlined]{algorithm2e}
\usepackage{booktabs}
\usepackage{arydshln}
\usepackage{pifont}
\usepackage{makecell}
\usepackage{orcidlink}

\usepackage{hyperref}
\hypersetup{
  colorlinks=true,
  linkcolor=blue,
  citecolor=blue,
  urlcolor=black
}

\newcommand{\shl}{\mathbin{\text{\texttt{<<}}}} 
\newcommand{\shr}{\mathbin{\text{\texttt{>>}}}} 

\begin{document}

\title{Super-LIO: A Robust and Efficient LiDAR-Inertial Odometry System with a Compact Mapping Strategy}

\author{
Liansheng Wang$^{1}$\,\orcidlink{0000-0003-0387-2296},
Xinke Zhang$^{1}$\,\orcidlink{0000-0001-6440-5240},
Chenhui Li$^{2}$,
Dongjiao He$^{3}$\,\orcidlink{0000-0003-0131-061X},
Yihan Pan$^{1}$\,\orcidlink{0009-0005-5095-4739},
Jianjun Yi$^{1}$\,\orcidlink{0000-0003-0899-177X}
\thanks{Received September~1,~2025; Revised November~24,~2025; Accepted December~19,~2025.
This paper was recommended for publication by the Associate Editor, Prof.~Sven Behnke, upon evaluation of the reviewers’ comments.
This work was supported by the Special Fund Technology Innovation Support Project of Shanghai under Grant HCXBCY-2023-046,
and by the Pioneer and Leading Goose R\&D Program of Zhejiang Province, No.~2025C02G5061814.
(Corresponding author: Jianjun Yi.)}
\thanks{$^{1}$Liansheng Wang, Xinke Zhang, Yihan Pan, and Jianjun Yi are with the
Department of Mechanical Engineering, East China University of Science and Technology,
Shanghai, China
(e-mail: lswang@mail.ecust.edu.cn; jjyi@ecust.edu.cn).}
\thanks{$^{2}$Chenhui Li is with the Shanghai Artificial Intelligence Laboratory,
Shanghai, China (e-mail: lichkermit@163.com).}
\thanks{$^{3}$Dongjiao He is with the University of Hong Kong, Hong Kong, China
(e-mail: hdj65822@connect.hku.hk).}
\thanks{Digital Object Identifier (DOI): see top of this page.}
}

\markboth{IEEE Robotics and Automation Letters. Preprint Version. December, 2025}
{Wang \MakeLowercase{\textit{et al.}}: Super-LIO: A Robust and Efficient LiDAR-Inertial Odometry System with a Compact Mapping Strategy}

\maketitle
\begin{abstract}
LiDAR-Inertial Odometry (LIO) is a foundational technique for autonomous systems, yet its deployment on resource-constrained platforms remains challenging due to computational and memory limitations. We propose Super-LIO, a robust LIO system that demands both high performance and accuracy, ideal for applications such as aerial robots and mobile autonomous systems. At the core of Super-LIO is a compact octo-voxel-based map structure, termed \textbf{OctVox}, that limits each voxel to eight subvoxel representatives, enabling strict point density control and incremental denoising during map updates. This design enables a simple yet efficient and accurate map structure, which can be easily integrated into existing LIO frameworks. Additionally, Super-LIO designs a heuristic-guided KNN strategy (HKNN) that accelerates the correspondence search by leveraging spatial locality, further reducing runtime overhead. We evaluated the proposed system using four publicly available datasets and several self-collected datasets, totaling more than 30 sequences. Extensive testing on both X86 and ARM platforms confirms that Super-LIO offers superior efficiency and robustness, while maintaining competitive accuracy. Super-LIO processes each frame approximately 73\% faster than SOTA, while consuming less CPU resources. The system is fully open-source and compatible with a wide range of LiDAR sensors and computing platforms. The implementation is available at: \url{https://github.com/Liansheng-Wang/Super-LIO.git}.
\end{abstract}

\section{Introduction}\label{sec1}

\begin{figure}[htbp]
\centering
\includegraphics[width=\columnwidth]{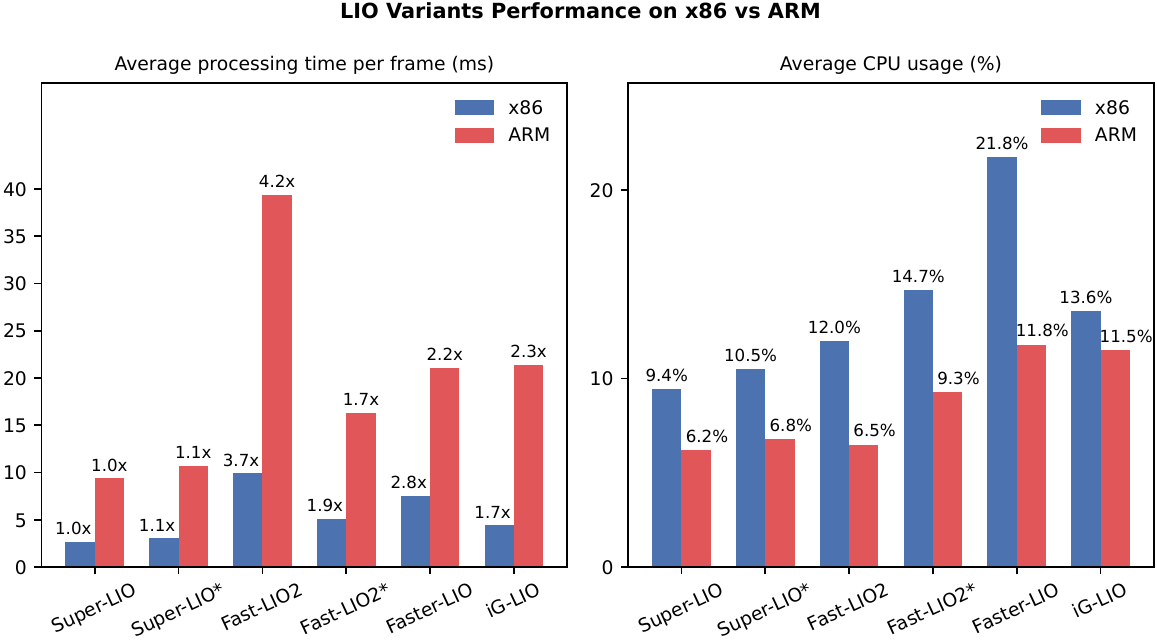}
\caption{Performance comparison of different LIO methods on X86 and ARM platforms. The left plot shows the average processing time per LiDAR frame, annotated relative to Super-LIO (1x) on each platform, while the right plot shows runtime CPU utilization. Note that the X86 experiments were conducted with 5x playback speed, which leads to higher CPU usage compared to ARM.}
\label{fig:perf}
\end{figure}

\IEEEPARstart{L}{iDAR}-Inertial Odometry (LIO) is a core module in autonomous navigation systems, fusing LiDAR geometry with inertial motion priors for accurate and robust pose estimation. It has been widely deployed in mobile robots, UAVs, and autonomous driving~\cite{AR_ugv,super,AotoDrive}. As robotic tasks grow in complexity, LIO increasingly operates as a foundational layer for higher-level autonomy, such as exploration~\cite{epic}, language-driven navigation~\cite{openfly}, and multi-agent coordination~\cite{uav_swarm}, where limited onboard computational resources must be shared among multiple modules. This competition for resources, especially on embedded or low-power platforms, makes it challenging for LIO to maintain real-time performance and high accuracy.

Early LiDAR-based Simultaneous Localization and Mapping (SLAM) systems, such as LOAM~\cite{LOAM} and LeGO-LOAM~\cite{lego}, decomposed the pipeline into feature extraction, scan-to-map registration, and map maintenance, establishing the foundation for LIO frameworks. LIO-SAM~\cite{lio_sam} further incorporates a pose-graph optimization backend to achieve loop closure and maintain global consistency. Later systems such as HBA~\cite{hba} and Voxel-SLAM~\cite{Voxel-slam} employ hierarchical optimization combined with bundle adjustment to perform global map refinement after the front-end odometry stage, further improving mapping quality. While these approaches achieve higher map fidelity, their optimization backends introduce additional computational stages, leading to higher per-frame latency and reducing their suitability for latency-sensitive deployment on resource-constrained platforms~\cite{fast_lio2, PointLIO, faster_lio}.

Filtering-based LIOs offer lightweight estimation and minimal latency, where the dominant computational cost lies in point-cloud registration due to extensive nearest-neighbor search. This cost and its consistency are largely determined by the underlying map structure, which governs both the number and spatial locality of candidate neighbors. Incrementally updatable maps such as the iKD-Tree~\cite{fast_lio2} and hash-voxel grids like iVox~\cite{faster_lio} accelerate point insertion and neighbor queries, but their runtime can still vary with factors such as local point density or voxel discretization. KISS-ICP~\cite{Kiss-icp} adopts a similar hash-voxel design that limits stored points per voxel to control memory usage. Although such structures improve efficiency, they rely on voxel-level thresholds rather than explicitly enforcing spatially uniform point distribution.

To overcome the density-related variability of existing voxel maps, we introduce \textbf{OctVox}, a compact hash-based octo-voxel representation that explicitly regularizes spatial density. Each voxel is subdivided into eight subvoxels, each maintaining an incrementally averaged representative point. This design provides bounded storage, uniform coverage, and a compact geometric representation, allowing a smaller yet well-structured point set to support accurate odometry while reducing computational and memory costs.

While OctVox stabilizes the map structure, achieving high-quality correspondences still requires searching beyond the small fixed neighborhoods used in prior voxel-based methods. Voxel-based neighbor queries~\cite{faster_lio, ig_lio} incur rapidly increasing cost as the search range expands. Building upon the ordered subvoxel layout of OctVox, we develop a heuristic-guided K-nearest-neighbor strategy (\textbf{HKNN}) that efficiently searches larger candidate regions with early termination, improving neighbor quality without incurring excessive overhead.

We integrate OctVox and HKNN into a unified filtering-based LIO framework, \textbf{Super-LIO}. Extensive experiments on four public datasets and multiple self-collected sequences across x86 and ARM platforms show that Super-LIO achieves accuracy comparable to state-of-the-art methods while reducing runtime and CPU usage (see Fig.~\ref{fig:perf}). The main contributions are summarized as follows:

\begin{itemize}
\item We propose a compact hash-based octo-voxel map representation (OctVox) that explicitly regularizes spatial density, achieving uniform coverage and noise suppression.
\item We design a heuristic-guided KNN strategy (HKNN) that leverages the structured subvoxel layout of OctVox to improve neighbor quality and search efficiency.
\item We integrate OctVox and HKNN into a unified filtering-based LIO system (Super-LIO) and validate it on diverse public and self-collected datasets across both x86 and ARM platforms, showing accuracy comparable to state-of-the-art methods with lower runtime and CPU usage.
\item We release the complete system to benefit the research community.
\end{itemize}

\section{Related Work}\label{sec2}
This section reviews prior LIO systems with a focus on map structures and correspondence search strategies, which dominate the computational cost in real-time deployment.

\textbf{Map structures and efficiency.} 
Early LIO systems such as LIO-SAM~\cite{lio_sam} and FAST-LIO~\cite{fast_lio} organize local maps using KD-Trees for correspondence search, whose maintenance cost becomes non-negligible due to frequent rebuilding in long-term incremental mapping.
FAST-LIO2~\cite{fast_lio2} addresses this limitation by introducing the iKD-Tree, enabling efficient online updates without full reconstruction.
Building upon the iKD-Tree framework, Point-LIO~\cite{PointLIO} further adopts point-wise state updates to increase update bandwidth and improve robustness under high-dynamic motion.

Faster-LIO~\cite{faster_lio} replaces tree-based indexing with a sparse hash voxel map, reducing insertion overhead and yielding more predictable access patterns.
However, the number of points stored per voxel is not explicitly controlled, causing correspondence search cost to depend on local point density and potentially increase in dense regions.
KISS-ICP~\cite{Kiss-icp} adopts a similar hash-voxel structure but caps the number of points per voxel to bound indexing cost, while still lacking explicit spatial density regulation.
Adaptive-LIO~\cite{adaptive-lio} adjusts voxel resolution according to scene openness, at the expense of increased structural complexity and higher neighborhood query cost.

Voxel-based statistical methods offer an alternative representation by replacing raw points with compact statistical models
VGICP~\cite{vgicp} estimates voxel-wise means and covariances to replace explicit nearest-neighbor queries with probabilistic distance evaluations. 
iG-LIO~\cite{ig_lio} builds upon this idea~\cite{vgicp} and integrates it into a tightly coupled LIO framework, improving overall efficiency. 
However, its statistical modeling requires environment-specific parameter tuning, and its computational demand is relatively high on embedded processors.
GLIM~\cite{GLIM} adopts a similar formulation but relies on GPU acceleration, which requires additional hardware support.

Octrees~\cite{ufomap, slict, i-octree} reduce memory through hierarchical subdivision, but high-resolution settings lead to deeper tree structures and longer traversal paths during neighbor lookup in high-rate LIO. MGM-LIO~\cite{MGM-LIO} employs multi-scale Gaussian maps, but its mapping and update procedures remain computationally heavy for real-time use on mobile robots. 
Recent Gaussian-splatting-based LiDAR mapping methods~\cite{Liv-GS, GS-LIVO} rely on GPU acceleration, making them less suited for lightweight platforms.

Across these designs, spatial density is generally unmanaged, or addressed indirectly through statistical modeling.
To address this, we introduce OctVox, which enforces subvoxel-level density regularization and provides a compact, structured representation for efficient mapping.

\textbf{KNN strategies and complexity.}
In scan-to-map registration, correspondence estimation relies on repeated nearest-neighbor queries and constitutes a major runtime bottleneck.

KD-tree-based methods~\cite{fast_lio2} perform KNN search through recursive space partitioning, whose backtracking behavior makes the search cost sensitive to point distribution and local geometry.

Hash-voxel-based approaches such as iVox~\cite{faster_lio} achieve constant-time voxel access via hashing, but the KNN cost depends on the number and spatial distribution of points within the queried voxels.
To bound this cost, searches are typically restricted to fixed neighboring voxel stencils; however, due to spatial discretization, true nearest neighbors may fall outside the stencil, leading to a resolution- and density-dependent trade-off between efficiency and correspondence quality.
Methods such as iG-LIO~\cite{ig_lio} and MGM-LIO~\cite{MGM-LIO} follow similar designs with per-voxel statistical models, but still rely on fixed voxel neighborhoods and incur additional probabilistic evaluation cost.

Building on the structured subvoxel layout of OctVox, we develop an HKNN strategy that leverages spatial priors to expand the effective search region while avoiding unnecessary point evaluations, enabling higher-quality correspondences at lower computational cost.

\begin{figure*}[htbp]
\centering
    \includegraphics[width=.95\textwidth]{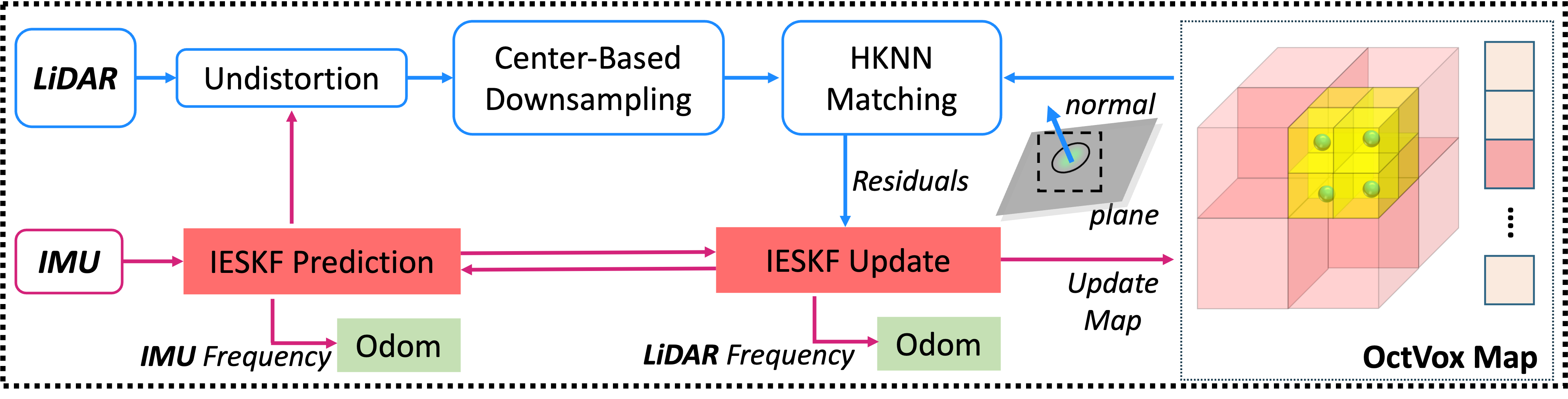}
    \caption{Overview of Super-LIO. An IESKF fuses IMU and LiDAR data: IMU measurements propagate the state at the IMU rate; at each LiDAR frame, points are de-skewed using the IMU state, center-based downsampled, and, via a heuristic-guided KNN (HKNN) in the OctVox Map, nearest neighbors are retrieved to construct point-to-plane residuals. These residuals feed the IESKF observation update. The map stores up to eight subvoxel representatives per voxel and is updated online to ensure efficient and accurate correspondences.}
    \label{fig:framework}
\end{figure*}

\section{Method}\label{sec:method}
\subsection{System Overview}

As shown in Fig.~\ref{fig:framework}, Super-LIO is a tightly coupled LiDAR-Inertial Odometry that follows an iterated error-state Kalman filter (IESKF) formulation~\cite{fast_lio}, while introducing two efficiency-centric modules: (i) OctVox, a sparse hash map that stores at most eight incrementally averaged representatives per voxel for explicit density control and incremental denoising (Sec.~\ref{sec:subvoxel}); and (ii) HKNN, a correspondence module that uses precomputed traversal lists and subvoxel symmetry to accelerate neighbor search within a small local neighborhood (Sec.~\ref{sec:hknn}). After IMU-aided de-skewing, each scan is downsampled via a center-based selection strategy that better preserves original measurements than standard voxel filters; the selected points are then used both to form scan-to-map point-to-plane residuals for IESKF and to update the OctVox map.

\subsection{State and Notation}\label{sec:state}
Let $\mathcal{G}$, $\mathcal{L}$, and $\mathcal{I}$ denote the Global, LiDAR, and IMU frames, respectively. The LiDAR-IMU extrinsic transform $\mathbf{T}_{\mathcal{L}}^{\mathcal{I}}=(\mathbf{R}_{\mathcal{L}}^{\mathcal{I}},\mathbf{p}_{\mathcal{L}}^{\mathcal{I}})$ is assumed to be well calibrated and fixed. The system state at time $t$ is defined as Eq.\eqref{eq_state}.
\begin{equation}
    \mathbf{X}_t \triangleq 
    \begin{bmatrix}
        \mathbf{R}_{\mathcal{I}}^{\mathcal{G}},\ \mathbf{p}_{\mathcal{I}}^{\mathcal{G}},\ \mathbf{v}_{\mathcal{I}}^{\mathcal{G}},\ 
        \mathbf{b}_a,\ \mathbf{b}_g,\ \mathbf{g}^{\mathcal{G}}
    \end{bmatrix}^{\!\top}
    \in SO(3)\times\mathbb{R}^{15}.
\label{eq_state}
\end{equation}

We denote estimates with a hat $\hat{(\cdot)}$ and use superscripts $\hat{(\cdot)}^{-}$/$\hat{(\cdot)}^{+}$ for priori and posteriori; unless stated otherwise, a bare symbol (e.g., $\mathbf{X}_t$) denotes ground truth.

IMU measurements $\{\mathbf{a}_i,\bm{\omega}_i\}$ drive propagation at the IMU rate. We use midpoint integration by averaging adjacent samples to reduce discretization error. For each scan anchored at $t_k$, raw LiDAR points are first expressed and de-skewed into the IMU frame at the anchor time $t_k$:
\begin{equation}
\label{eq:deskew_to_I}
\bm{p}^{\mathcal{I}_k}_i = \mathbf{R}_{\mathcal{I}_i}^{\mathcal{I}_k}\!\left(\mathbf{R}_{\mathcal{L}}^{\mathcal{I}}\bm{p}^{\mathcal{L}}_i+\mathbf{p}_{\mathcal{L}}^{\mathcal{I}}\right) + \mathbf{p}_{\mathcal{I}_i}^{\mathcal{I}_k}, \ t_i\!\in\![t_{k-1},t_k],
\end{equation}
where $(\mathbf{R}_{\mathcal{I}_i}^{\mathcal{I}_k},\,\mathbf{p}_{\mathcal{I}_i}^{\mathcal{I}_k})$ denotes the relative IMU motion from $t_i$ to $t_k$ obtained by propagation and subsequent interpolation between IMU samples, for example using a constant-velocity model~\cite{LOAM}, a constant-acceleration model~\cite{fast_lio2}, or a higher-order scheme~\cite{dlio}. De-skewed LiDAR points are expressed in the IMU frame $\mathcal{I}_k$ and transformed to the global frame via Eq.~\eqref{eq:I_to_G} when needed for association or map updates.

\begin{equation}
\label{eq:I_to_G}
\bm{p}^{\mathcal{G}}_i = \mathbf{R}_{\mathcal{I}}^{\mathcal{G}}(t_k)\,\bm{p}^{\mathcal{I}_k}_i + \mathbf{p}_{\mathcal{I}}^{\mathcal{G}}(t_k).
\end{equation}

\subsection{Octo-Voxel-Based Map (OctVox)} \label{sec:subvoxel}

\textbf{Data structure.} We employ a hashed voxel grid with a least-recently-used (LRU) policy for voxel caching~\cite{faster_lio}. The map is defined as $\mathcal{M} = \{ \mathcal{V}_i \mid i \in \mathbb{N} \}$, i.e., a collection of voxels. Each voxel $\mathcal{V}_i$ of edge length $r_v$ is subdivided into $2\times2\times2$ subvoxels $\mathcal{V}_{i,s}$ of edge length $r_s=\tfrac{1}{2}r_v$. A voxel allocates eight contiguous entries, each storing a representative $\boldsymbol{\mu}_{i,s}\in\mathbb{R}^3$ and a counter $n_{i,s}$. The structure can be expressed as
\[
\mathcal{V}_i = \{ \mathcal{V}_{i, s} \mid s \in \{0,\dots,7\} \}, \ \ \mathcal{V}_{i, s} = (\boldsymbol{\mu}_{i,s},n_{i,s}).
\]
This contiguous layout enforces an explicit density cap of eight representatives per voxel and achieves noise suppression in the global frame through incremental averaging within each subvoxel. The hash table uses Robin Hood hashing with open addressing~\cite{celis1985robin} and is implemented with \texttt{tsl::robin\_map}~\cite{tsl_robin_map}, improving cache locality and providing near-constant lookup time. When the hash table reaches its maximum voxel capacity, the LRU policy evicts the least-recently-accessed voxels.

\textbf{Voxel indexing and update. }
Given a point $\hat{\bm{p}}^\mathcal{G}\in\mathbb{R}^3$ in the world frame, we quantize it to subvoxel resolution using Eq.\eqref{eq:subvoxel_indexing} to obtain its parent voxel key $\mathbf{k}\in\mathbb{Z}^3$ and the local linear subvoxel index $s$.
\begin{equation}
\label{eq:subvoxel_indexing}
\begin{aligned}
&\mathbf{k}^{\text{sub}} = \Big\lfloor \tfrac{\hat{\bm{p}}^\mathcal{G}}{r_{\text{s}}} \Big\rfloor,  \quad
\mathbf{k} = \mathbf{k}^{\text{sub}} \shr 1, \\
& b_x = {k}^{\text{sub}}_x \,\&\, 1, \ \  b_y = {k}^{\text{sub}}_y \,\&\, 1, \ \  b_z = {k}^{\text{sub}}_z \,\&\, 1\\
&s = b_x \;|\;\big(b_y\,\shl 1\big) \;|\;\big(b_z\,\shl 2\big)
\end{aligned}
\end{equation}
Here, $\&$ and $|$ denote bitwise AND/OR, while $\shl$ and $\shr$ denote left and right shifts, respectively. Equations~\eqref{eq:subvoxel_indexing} therefore provide both the voxel key $\mathbf{k}$ and the subvoxel index $s$ in $O(1)$ time using bitwise operations.

After obtaining $(\mathbf{k}, s)$, the corresponding subvoxel entry $(\boldsymbol{\mu}_s, n_s)$ is retrieved from the hash table. If uninitialized, it is set to $(\hat{\bm{p}}^{\mathcal{G}}, 1)$; otherwise, when $\lVert \hat{\bm{p}}^{\mathcal{G}}-\boldsymbol{\mu}_s \rVert_2 \le \tau_{\text{merge}}$ and $n_s \le n_{\text{max}}$, an incremental mean update is applied as Eq.~\eqref{eq:point_update}. This corresponds to an unbiased incremental mean update, with the estimation variance asymptotically decreasing as $1/n_s$.

\begin{equation}
\label{eq:point_update}
\boldsymbol{\mu}_s \leftarrow \boldsymbol{\mu}_s + \tfrac{1}{n_s+1}\big(\hat{\bm{p}}^\mathcal{G}-\boldsymbol{\mu}_s\big),
\quad n_s \leftarrow n_s+1,
\end{equation}

This procedure resembles online voxel downsampling in the global frame. It enforces bounded density by maintaining at most one representative per subvoxel, while progressively suppressing measurement noise through averaging. While conceptually simple, the results in Sec.~\ref{sec:exp_accu} indicate that this design maintains accuracy.

\subsection{Heuristic-Guided KNN Search}\label{sec:hknn}
A major bottleneck in LiDAR-Inertial Odometry is the repeated KNN search for scan-to-map residuals. Even with OctVox map that cap per-voxel density, a naive search must examine all representatives within a ball of radius $R$. Small $R$ often yields unstable neighbors in sparse areas, whereas larger $R$ improves quality but requires visiting $\Theta((R/r_v)^3)$ voxels, so the candidate set grows cubically with $R$. Tree-based indices (e.g., KD-Tree) scale as $\mathcal{O}(\log N)$ with map size, but lack the constant-time locality of voxel hashing. To address this, we design an HKNN that groups candidate subvoxels by distance and exploits subvoxel symmetry to unify traversal, enabling efficient and robust correspondence search.

\begin{figure}[htbp]
\centering
\includegraphics[width=.95\columnwidth]{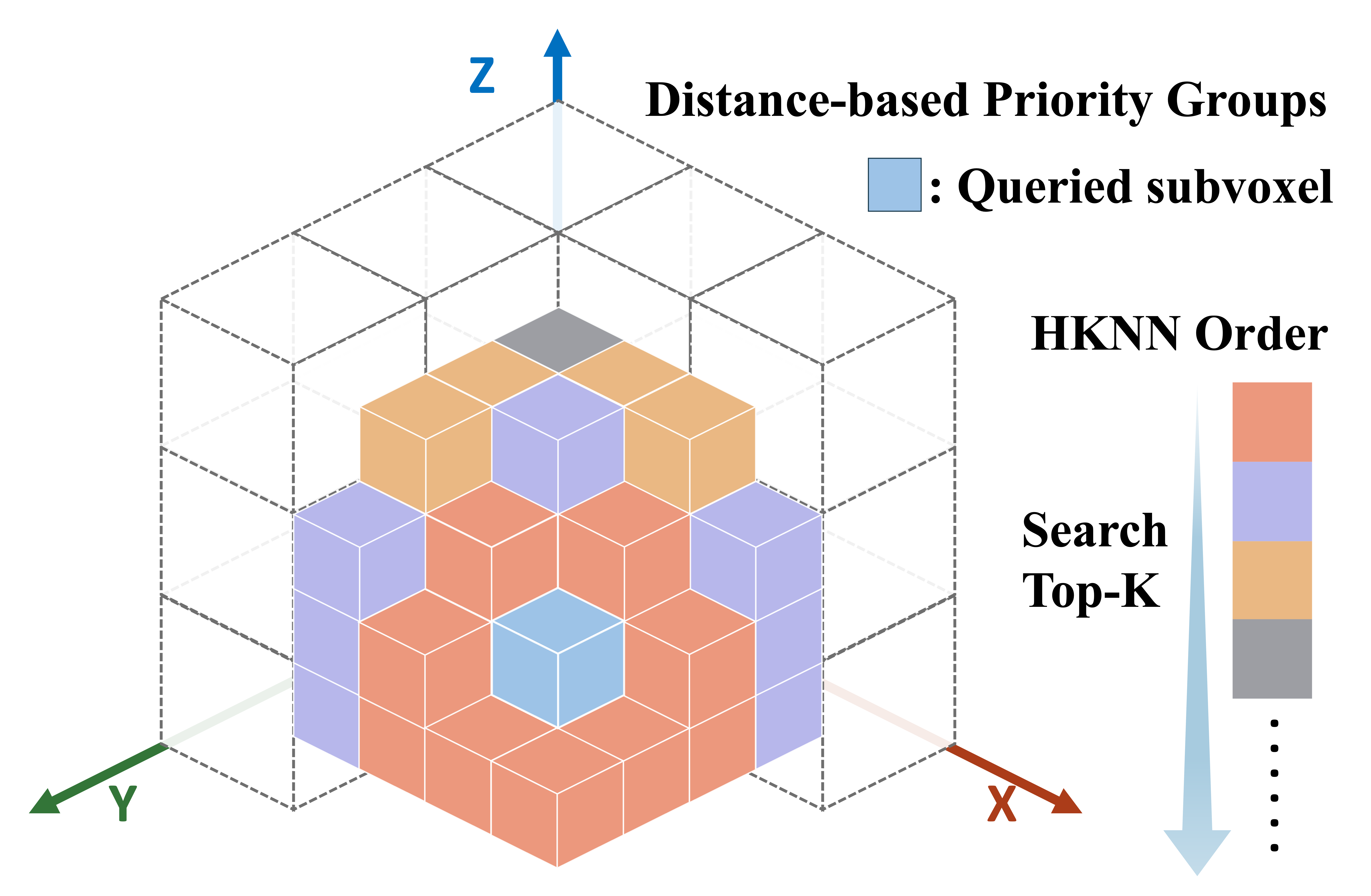}
\caption{Illustration of the HKNN candidate selection process based on subvoxel proximity. A query subvoxel (blue) is surrounded by neighboring subvoxels at increasing distances, which are color-coded by the minimum Euclidean distance between their nearest vertices. These candidates are pre-grouped by geometric proximity; the groups are traversed sequentially in order of increasing distance until $K$ neighbors are found and no subsequent group can yield a closer neighbor, enabling efficient Top-$K$ retrieval without per-query sorting.}
\label{fig:hknn_groups}
\end{figure}

\textbf{HKNN list generation.} 
As shown in Fig.\ref{fig:hknn_groups}, the HKNN list is defined as an ordered sequence of subvoxel groups, each group contains candidate subvoxels clustered by their spatial distance from the query subvoxel. The list is generated once during initialization from the maximum search radius and voxel resolution and remains fixed throughout operation. In contrast to conventional neighborhood definitions\cite{faster_lio,ig_lio} that require exhaustive traversal of a fixed set, the HKNN list enables group-wise traversal with early termination.

We define each subvoxel as a cube with eight vertices $\{C_i\}_{i=1}^8$. 
Given two subvoxels $A$ and $B$, their distance is defined as the minimum Euclidean distance between their vertex sets, as given in Eq.~\eqref{eq:subvoxel_dist}. By definition, $\mathrm{dis}(A,A)=0$ and $\mathrm{dis}(A,B)=\mathrm{dis}(B,A)$.

\begin{equation}
\label{eq:subvoxel_dist}
\mathrm{dis}(A,B) = \min_{C_i \in A,\, C_j \in B} \| C_i - C_j \|_2.
\end{equation}

Let $\mathbf{K}=\{\mathbf{k}_i\}$ denote the set of voxel indices within the maximum search radius $R_{\max}$, where each index $\mathbf{k}_i$ is defined relative to the query voxel center at the origin,  $\mathbf{k}_0 = (0,0,0)$.
The corresponding set of subvoxels is given by $\mathbf{S} \;=\; \mathbf{K} \times \{0,\ldots,7\}$, and each subvoxel element is indexed by $(\mathbf{k}_i, s_j)$ and denoted as $\mathbf{S}_{i,j}$.

Due to the discrete nature of the voxel grid, the distances between the reference subvoxel $\mathbf{S}_{0,0}$ and all subvoxels $\mathbf{S}_{i,j}\in\mathbf{S}$, computed by Eq.~\eqref{eq:subvoxel_dist}, constitute a finite enumerated set of attainable distances.
We denote this set by $\mathcal{D}=\{d_m \mid m=0,\dots,M\}$, which is arranged in ascending order.
Subvoxels at the same distance are then grouped into a distance-equivalence set, as defined in Eq.\eqref{eq:group_def}.
\begin{equation}
\label{eq:group_def}
\mathcal{H}_m \;=\; \big\{\, \mathbf{S}_{i,j}\in\mathbf{S} \;:\; \mathrm{dis}\big(\mathbf{S}_{0,0},\mathbf{S}_{i,j}\big)=d_m \,\big\}
\end{equation}

Concatenating the distance-equivalence groups yields the ordered traversal list for the reference subvoxel $\mathbf{S}_{0,0}$ of the query voxel, as defined in Eq.~\eqref{eq:traversal_concat}.

\begin{equation}
\label{eq:traversal_concat}
\mathcal{T}(\mathbf{S}_{0,0}) \;=\; \big[\, \mathcal{H}_0 \,\Vert\, \mathcal{H}_1 \,\Vert\, \cdots \,\Vert\, \mathcal{H}_M \,\big]
\end{equation}

Traversal lists for the remaining seven subvoxels are obtained by reflecting the reference list $\mathcal{T}(\mathbf{S}_{0,0})$ across the three coordinate axes. 
For each entry $(k_i^x,k_i^y,k_i^z,s_j)$ in $\mathcal{T}(\mathbf{S}_{0,0})$,
the corresponding voxel and subvoxel indices $(\mathbf{k}'_i, s'_j)$ are computed as Eq.~\eqref{eq:symmetry}, where $s$ is the local subvoxel index, $(b_x,b_y,b_z)$ are the bit variables of $s$ defined in Eq.~\eqref{eq:subvoxel_indexing}, and $\oplus$ denotes bitwise XOR. According to Eq.~\eqref{eq:symmetry}, the symmetry operation leaves $\mathcal{T}(\mathbf{S}_{0,0})$ unchanged.

\begin{equation}
\label{eq:symmetry}
\mathbf{k}'_i = \big((-1)^{b_x} k_i^x,\; (-1)^{b_y} k_i^y,\; (-1)^{b_z} k_i^z \big), \quad s'_j = s \oplus s_j ,
\end{equation}

\textbf{Heuristic-Guided KNN.}
The HKNN search exploits the precomputed priority list in Eq.~\eqref{eq:traversal_concat} to accelerate scan-to-map correspondence search.
Given a query point, we obtain a global prior via Eq.~\eqref{eq:I_to_G} and restrict the correspondence search in $\mathcal{M}$ to a fixed neighborhood of radius $R$, returning the $K$ nearest neighbors.
The search traverses candidate groups in the precomputed priority order; occupied subvoxels are evaluated using exact Euclidean distances to their stored representatives, while a bounded max-heap maintains the current top-$K$ neighbors.
Upon completing a group, if the heap is full and its worst distance $r_K$ is strictly less than the lower bound of the next group, the search terminates.
The lower-bound property of the grouping certifies that any unvisited candidate is no closer than that threshold and thus cannot reduce $r_K$.
Together, these properties produce exact top-$K$ neighbors with geometric completeness while substantially reducing candidate evaluations.

The detailed procedure is summarized in Algorithm~\ref{alg:hknn_final}, where $|\mathbf{H}|$ and $\mathrm{maxdist}(\mathbf{H})$ denote the number of elements in the max-heap $\mathbf{H}$ and the largest distance among them, respectively. $\mathcal{V}(\mathbf{k}_g, s')$ denotes the unique subvoxel identified by the global voxel index $\mathbf{k}_g$ and the subvoxel index $s'$.

\begin{algorithm}[t]
\caption{Heuristic-Guided KNN (HKNN)}
\label{alg:hknn_final}
\KwIn{Query point $\bm{p}^{\mathcal{I}}$ (IMU frame); prior state $\hat{\mathbf{X}}^{-}$; voxel map $\mathcal{M}$; canonical traversal list $\mathcal{T}(\mathbf{S}_{0,0})$ Per Eq.~\eqref{eq:traversal_concat}; group lower bounds $\mathcal{D}=\{d_m\}_{m=0}^{M}$; fixed radius $R \le R_{max} $; target $K$}
\KwOut{Top-$K$ nearest neighbors $\mathcal{N}_K$}

\BlankLine
Transform $\bm{p}^{\mathcal{I}}$ to $\hat{\bm{p}}^{\mathcal{G}}$ using $\hat{\mathbf{X}}^{-}$ (Eq.~\eqref{eq:I_to_G}).

Compute the global voxel index $\mathbf{k}_p$, obtain $(b_x,b_y,b_z)$ and hence $s_p$ via Eq.~\eqref{eq:subvoxel_indexing}.

Let $m^\star = \max\{\, m \mid d_m \le R \,\}$; 

Set $\mathcal{T}^\star = [\,\mathcal{H}_0 \,\Vert\, \cdots \,\Vert\, \mathcal{H}_{m^\star}\,]$ and $\mathcal{D}^\star = \{d_0,\dots,d_{m^\star}\}$.

Initialize a bounded max-heap $\mathbf{H}$ with capacity $K$.

\BlankLine

\For{$i \in [0, m^\star]$}{
  \lIf{$|\mathbf{H}|=K$ \textbf{and} $d_i > \mathrm{maxdist}(\mathbf{H})$}{\textbf{break}}
  
  \ForEach{$(k_x,k_y,k_z,\ s)\ \in\ \mathcal{H}_i$}{
    \tcp{Eq.~\eqref{eq:symmetry}: octant reflection}
    
    $\sigma_x \gets (-1)^{b_x},\ \sigma_y \gets (-1)^{b_y},\ \sigma_z \gets (-1)^{b_z}$
    
    $\mathbf{k}' \gets (\sigma_x k_x,\ \sigma_y k_y,\ \sigma_z k_z)$
    
    $\mathbf{k}_g \gets \mathbf{k}_p + \mathbf{k}'$ \ and \ $s' \gets s \oplus s_p$
    
    $\boldsymbol{\mu} \gets \mathcal{V}(\mathbf{k}_g, s') \gets (\mathbf{k}_g, s')$
    
    $\delta \gets \big\|\hat{\bm{p}}^{\mathcal{G}} - \boldsymbol{\mu}\big\|_2$
    
    \lIf{$\delta > R$}{\textbf{continue}}
    
    \lIf{$|\mathbf{H}|<K$}{push $(\delta,\boldsymbol{\mu})$ into $\mathbf{H}$; \textbf{continue}}
    
    \If{$\delta < \mathrm{maxdist}(\mathbf{H})$}{
        pop worst from $\mathbf{H}$; push $(\delta,\boldsymbol{\mu})$ into $\mathbf{H}$
      }
  }
}
$\mathcal{N}_K \gets$ extract elements of $\mathbf{H}$ in ascending order of $\delta$.

\end{algorithm}

\section{Experiment} \label{sec4}

In this section, we evaluate Super-LIO on both X86 and ARM platforms and across multiple datasets, comparing it against state-of-the-art lightweight LIO frameworks. The experiments cover trajectory accuracy, per-platform runtime, module-level time analysis, and system resource usage such as CPU load and memory consumption.

\subsection{Experimental Setup}\label{sec:exp_setup}

We compare \textbf{Super-LIO} against several lightweight LiDAR-Inertial Odometry baselines, including \textbf{FAST-LIO2}~\cite{fast_lio2}, \textbf{Faster-LIO}~\cite{faster_lio}, and \textbf{iG-LIO}~\cite{ig_lio}. 
To isolate the contribution of the HKNN strategy, we additionally implement an ablation variant, \textbf{Super-LIO*}, which is identical to Super-LIO but replaces HKNN with the 18-neighbor voxel search of Faster-LIO. 
We also provide a concurrency-optimized version of FAST-LIO2, denoted \textbf{FAST-LIO2*}, which preserves the original algorithm but replaces the OpenMP-based parallel implementation in the official release with Intel TBB, ensuring that its concurrency behavior matches that of Super-LIO for fair runtime and resource-usage evaluation.

For fairness, all methods use identical parameters: maximum iterations = 4, random downsampling rate = 3, and voxel-filter resolution = 0.5\,m. 
For voxel-based methods~\cite{faster_lio, ig_lio} including Super-LIO, the map voxel size is also set to 0.5\,m. 
Super-LIO uses an HKNN radius of $R_{\text{max}}=0.875\,\mathrm{m}$ and searches within a $7\times7\times7$ subvoxel neighborhood.
Except for dataset-specific LiDAR-IMU extrinsics, all parameters remain unchanged across experiments.

\begin{figure}[htbp]
  \centering
  \includegraphics[width=0.95\linewidth]{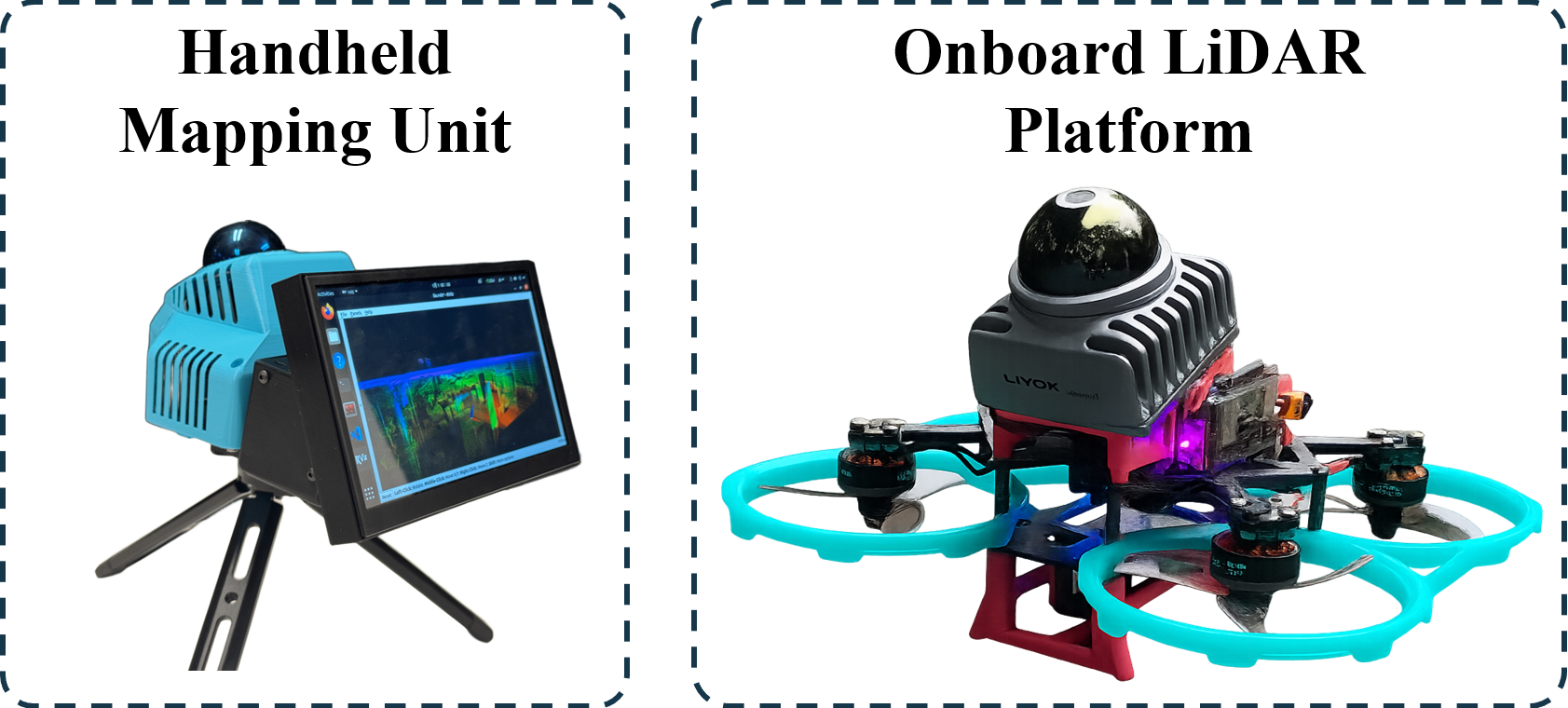}
  \caption{
  Experimental platforms used for dataset collection. 
  (\textbf{Left}) A handheld mapping unit equipped with a Livox-MID360 LiDAR, RGB camera, and an NVIDIA Orin NX embedded processor.
  (\textbf{Right}) A 220\,mm quadrotor UAV carrying the same LiDAR and embedded processor, used for autonomous flight experiments.
  }
  \label{fig:platforms}
\end{figure}

\begin{figure*}[htbp]
\centering
    \includegraphics[width=.99\textwidth]{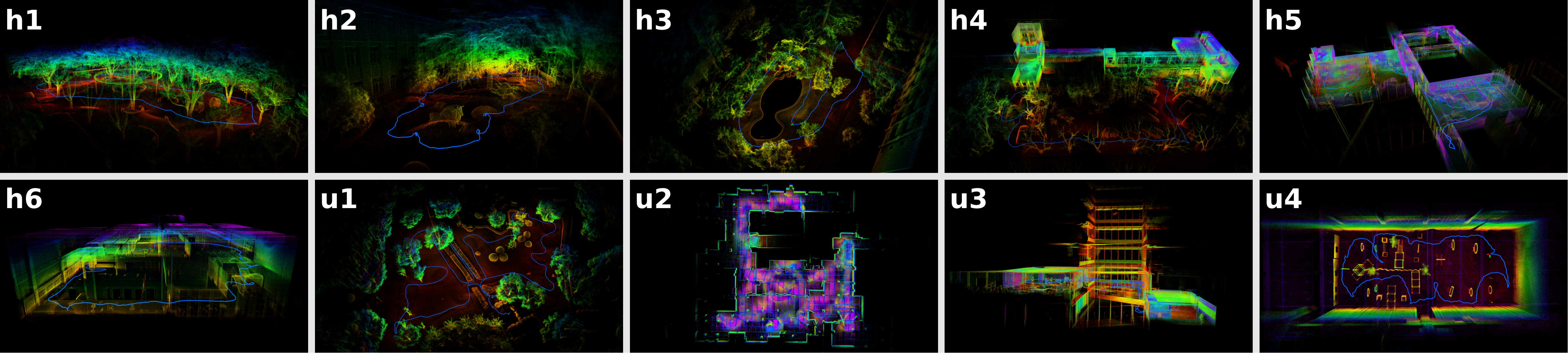}
    \caption{
    Point cloud maps produced by Super-LIO on ten self-collected evaluation sequences.  
    The sequences span diverse environments, including forests, parks, underground garages, and office buildings.  
    “h” denotes data collected with a handheld device, and “u” denotes data collected on a UAV platform.
    These varied scenarios and motion conditions are used to evaluate computational efficiency and robustness.}
    \label{fig:se_dataset}
\end{figure*}

Experiments are conducted on both public and private datasets. 
Public benchmarks include M2DGR~\cite{m2dgr}, NCLT~\cite{nclt}, MCD~\cite{MCD}, and NTU VIRAL~\cite{NTU}, all providing ground-truth trajectories. 
Our private dataset contains ten isequences covering diverse indoor and outdoor environments, used for runtime and robustness tests.
All data are replayed at $5\times$ speed on the X86 platform(AMD 5800H) to evaluate throughput and accuracy, and at $1\times$ speed on the NVIDIA Orin NX to evaluate real-time performance under resource constraints.
Examples of private data collection platforms and sequences are shown in Fig.~\ref{fig:platforms} and Fig.~\ref{fig:se_dataset}.

\subsection{Accuracy Evaluation}\label{sec:exp_accu}

\begin{table}[t]
  \centering
  \caption{RMSE (m) on Public Datasets (X86 @ 5x Speed)}
  \label{tab:accuracy}
  \setlength{\tabcolsep}{1pt}
  \begin{tabular*}{\columnwidth}{@{\extracolsep{\fill}}lcccccccc@{}}
    \toprule
    \midrule
    Seq. & {\scriptsize Super-LIO} & {\scriptsize Super-LIO*} & {\scriptsize FAST-LIO2} & {\scriptsize FAST-LIO2*} & {\scriptsize Faster-LIO} & {\scriptsize iG-LIO} & {\scriptsize Dist.(km)} \\
    \midrule
    m2s1  & 0.384 & 0.401 & 0.381 & \underline{0.325} & \textbf{0.323} & 0.328 & 0.752 \\
    m2s2  & \textbf{2.801} & 3.052 & \underline{2.862} & 2.891 & 3.109 & 2.968 & 1.483 \\
    m2s3  & \textbf{0.139} & \underline{0.148} & 0.198 & 0.195 & 0.167 & 0.201 & 0.424 \\
    m2s4  & 0.641 & 0.671 & \underline{0.455} & 0.470 & 0.601 & \textbf{0.429} & 0.840 \\
    m2s5  & 0.382 & 0.394 & 0.379 & \underline{0.369} & 0.394 & \textbf{0.274} & 0.421 \\
    m2h1  & \textbf{0.291} & \underline{0.294} & 0.303 & 0.302 & 0.298 & 0.305 & 0.214 \\
    m2h5  & \textbf{1.162} & \underline{1.166} & 1.198 & 1.197 & 1.194 & 1.170 & 0.285 \\
    m2d1  & 0.456 & 0.454 & 0.465 & 0.465 & \underline{0.450} & \textbf{0.442} & 0.143 \\
    m2d2  & \textbf{0.321} & \underline{0.339} & 0.364 & 0.364 & 0.341 & 0.371 & 0.057 \\
    nclt1 & \textbf{1.692} & 1.763 & \underline{1.719} & 1.730 & 2.022 & 2.475 & 7.582 \\
    nclt2 & \underline{1.296} & 1.316 & 1.419 & 1.481 & 1.325 & \textbf{1.170} & 3.186 \\
    nclt3 & \textbf{1.821} & 1.934 & 2.289 & 2.330 & 2.421 & \underline{1.876} & 6.121 \\
    mcd0  & \underline{0.541} & 0.634 & \textbf{0.484} & 0.572 & 0.574 & 1.289 & 3.197 \\
    mcd2  & \textbf{0.721} & \underline{0.790} & 1.085 & 1.413 & 0.951 & 1.692 & 1.784 \\
    mcd3  & \textbf{0.498} & 0.560 & 0.583 & 0.748 & \underline{0.557} & 0.921 & 1.459 \\
    mcd4  & \textbf{0.604} & \underline{0.629} & 0.964 & 1.117 & 0.695 & 0.752 & 2.421 \\
    eee1  & 0.119 & 0.123 & \textbf{0.079} & \underline{0.083} & 0.164 & \ding{55} & 0.261 \\
    nya1  & 0.069 & 0.078 & \textbf{0.063} & \underline{0.064} & 0.127 & 0.072 & 0.193 \\
    sbs1  & \textbf{0.086} & \underline{0.086} & 0.087 & 0.088 & 0.131 & 0.088 & 0.219 \\
    \midrule
    Avg   & \textbf{0.738} & \underline{0.781} & 0.809 & 0.853 & 0.834 & 0.934 & 1.634 \\
    \midrule
    \bottomrule
  \end{tabular*}
  \begin{flushleft}
  \footnotesize{
  The symbol \ding{55} denotes a failed run. Styling rule: per row, all minima are \textbf{bold}; second-smallest distinct values are \underline{underlined}. Ties share the same style.}
  \end{flushleft}
\end{table}

Accuracy is evaluated on public datasets with ground-truth trajectories, as introduced in Section~\ref{sec:exp_setup}. 
We use the Root Mean Square Error (RMSE) between estimated and ground-truth poses, computed by the Evo toolbox~\cite{evo}, as the evaluation metric.

For brevity, dataset names are simplified in Table~\ref{tab:accuracy}: for example, M2DGR~\cite{m2dgr} sequences are denoted by the prefix ``m2'' and the NTU VIRAL~\cite{NTU} prefix is omitted in the last three rows. The same naming rule applies in the following analysis.

The evaluated datasets cover a wide spectrum of LiDAR types, platforms, and scenarios. M2DGR~\cite{m2dgr} and NCLT~\cite{nclt} both employ a Velodyne HDL-32E spinning LiDAR. M2DGR consists of low-speed ground-robot runs, where we evaluate outdoor street and indoor corridor scenes. NCLT provides long-duration trajectories (up to 110 minutes and 7.58\,km), allowing us to assess accuracy under extended continuous operation. The MCD~\cite{MCD} dataset uses a Livox MID-70 solid-state LiDAR with a small FoV, recorded on a high-speed terrain vehicle; the tested sequences exceed 1\,km, with peak speed $\sim$10\,m/s and angular velocity $\sim$2.9\,rad/s, stressing both robustness and accuracy under fast motion. Finally, the NTU VIRAL~\cite{NTU} dataset is collected with a UAV equipped with an Ouster-16 LiDAR, yielding relatively sparse aerial point clouds; this setting is particularly challenging, and iG-LIO~\cite{ig_lio} fails on one sequence (eee1).

Table~\ref{tab:accuracy} reports the RMSE results across multiple sequences. Super-LIO achieves the best average accuracy among all tested methods, despite relying on a significantly more compact map representation. In several sequences, it even surpasses state-of-the-art baselines, illustrating that the OctVox map provides a reliable geometric representation. The ablation variant Super-LIO* shows a modest yet expected degradation, confirming that the proposed HKNN search improves the data association quality during scan-to-map alignment. By contrast, FAST-LIO2* exhibits a slight accuracy drop compared with FAST-LIO2~\cite{fast_lio2}, which is mainly attributed to more aggressive multi-threading that introduces small numerical inconsistencies in scan-to-map optimization.

\subsection{Efficiency Analysis}

This section evaluates the runtime performance and system-resource usage of all methods on both the X86 laptop and the embedded NVIDIA Orin NX. 
Experiments cover all public datasets as well as ten additional self-collected (prefix ``se'') Livox-MID-360 sequences.

Table~\ref{tab:x86_runtime} and Table~\ref{tab:arm_runtime} report the scene-level average per-frame processing time on the x86 laptop and the NVIDIA Orin NX, respectively.
For each scene, all available sequences are averaged to obtain a single representative runtime measurement. 
Across all public and self-collected scenes, Super-LIO achieves the lowest processing latency on both platforms, demonstrating a consistent efficiency gain across hardware architectures. 
As illustrated in Fig.~\ref{fig:perf}, it delivers \textbf{$3.7\times$} and \textbf{$4.2\times$} speedups over FAST-LIO2 on X86 and ARM, respectively, while also requiring fewer CPU resources. In contrast, iG-LIO performs competitively on x86 but incurs higher latency on ARM.

\begin{table}[t]
  \centering
  \caption{X86 per-frame runtime(ms), averaged across sequence groups.}
  \label{tab:x86_runtime}
  \footnotesize
  \setlength{\tabcolsep}{2pt}
  \resizebox{\columnwidth}{!}{
  \begin{tabular}{lcccccc}
    \toprule
    \midrule
      & Super-LIO & Super-LIO* & FAST-LIO2 & FAST-LIO2* & Faster-LIO & iG-LIO \\
    \midrule
    m2s  & \textbf{4.75} & \underline{5.40} & 19.41 & 8.85 & 14.82 & 7.79 \\
    m2h  & \textbf{2.70} & \underline{3.43} & 11.27 & 6.59 & 11.84 & 5.21 \\
    m2d  & \textbf{2.89} & \underline{3.29} & 11.02 & 6.47 & 9.29 & 4.70 \\
    nclt & \textbf{4.19} & \underline{5.14} & 13.53 & 7.32 & 9.90 & 5.77 \\
    mcd  & \textbf{1.73} & \underline{2.08} & 5.50 & 3.06 & 3.16 & 3.23 \\
    ntu  & \textbf{2.80} & \underline{3.15} & 8.95 & 4.39 & 8.32 & 3.72 \\
    se   & \textbf{1.88} & \underline{1.97} & 6.59 & 3.01 & 4.24 & 3.79 \\
    \midrule
    Avg  & \textbf{2.99} & \underline{3.49} & 10.90 & 5.67 & 8.80 & 4.89 \\
    \midrule
    \bottomrule
  \end{tabular}
  }
\end{table}

\begin{table}[t]
  \centering
  \caption{ARM per-frame runtime(ms), averaged across sequence groups.}
  \label{tab:arm_runtime}
  \footnotesize
  \setlength{\tabcolsep}{2pt}
  \resizebox{\columnwidth}{!}{
  \begin{tabular}{lcccccc}
    \toprule
    \midrule
      & Super-LIO & Super-LIO* & FAST-LIO2 & FAST-LIO2* & Faster-LIO & iG-LIO \\
    \midrule
    m2s  & \textbf{18.77} & \underline{19.84} & 84.74 & 28.63 & 38.81 & 40.47 \\
    m2h  & \textbf{9.87} & \underline{11.21} & 44.61 & 22.58 & 29.63 & 22.45 \\
    m2d  & \textbf{11.22} & \underline{12.08} & 48.16 & 22.51 & 28.97 & 24.60 \\
    nclt & \textbf{9.30} & \underline{11.01} & 36.87 & 16.57 & 19.79 & 17.02 \\
    mcd  & \textbf{6.03} & \underline{7.35} & 20.40 & 10.57 & 10.65 & 10.01 \\
    ntu  & \textbf{8.94} & \underline{9.84} & 29.06 & 12.82 & 15.25 & \ding{55} \\
    se   & \textbf{7.17} & \underline{8.29} & 25.52 & 9.64 & 12.64 & 14.89 \\
    \midrule
    Avg  & \textbf{10.47} & \underline{11.66} & 41.34 & 17.76 & 22.53 & 21.24 \\
    \midrule
    \bottomrule
  \end{tabular}
  }
\end{table}

\begin{table}[h]
  \centering
  \caption{Relative efficiency $\uparrow$ on X86 and ARM platforms.}
  \label{tab:re_results}
  \footnotesize
  \setlength{\tabcolsep}{1pt}
  \begin{tabular}{lcccccc}
    \toprule
     & Super-LIO & Super-LIO* & FAST-LIO2 & FAST-LIO2* & Faster-LIO & iG-LIO \\
    \midrule
    X86 & \textbf{3.98} & \underline{3.12} & 0.84 & 1.32 & 0.60 & 1.66 \\
    ARM & \textbf{1.71} & \underline{1.37} & 0.39 & 0.66 & 0.40 & 0.43 \\
    \bottomrule
  \end{tabular}
\end{table}

To complement the average processing time, we also report a composite relative efficiency metric that jointly accounts for runtime and CPU usage:
\[
\eta = \frac{1}{\left(\tfrac{1}{N}\sum t_i\right)\left(\tfrac{1}{N}\sum u_i\right)} .
\]
Here, $t_i$ is the per-frame processing time (ms) and $u_i$ is the normalized CPU utilization ($0$-$1$). As summarized in Table~\ref{tab:re_results}, this metric consistently highlights the strong efficiency of Super-LIO across both platforms.

\begin{figure}[th]
\centering
\includegraphics[width=\columnwidth]{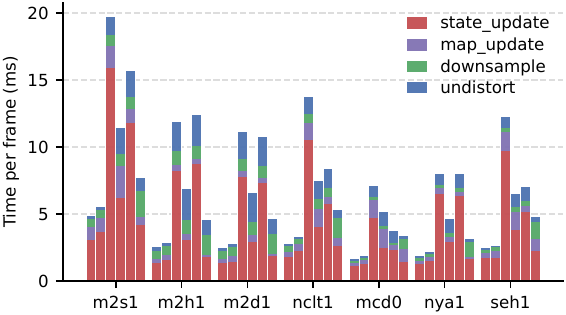}
\caption{Module-level timing analysis of per-frame processing for the evaluated 
LIO frameworks. Each bar shows the time spent in the four core stages of 
the pipeline: undistortion, downsample, state update, and map update. 
Within each scene, the methods appear from left to right as: Super-LIO, 
Super-LIO*, Fast-LIO2, Fast-LIO2*, Faster-LIO, and iG-LIO.}
\label{fig:module_time}
\end{figure}

\begin{figure}[th]
\centering
\includegraphics[width=\columnwidth]{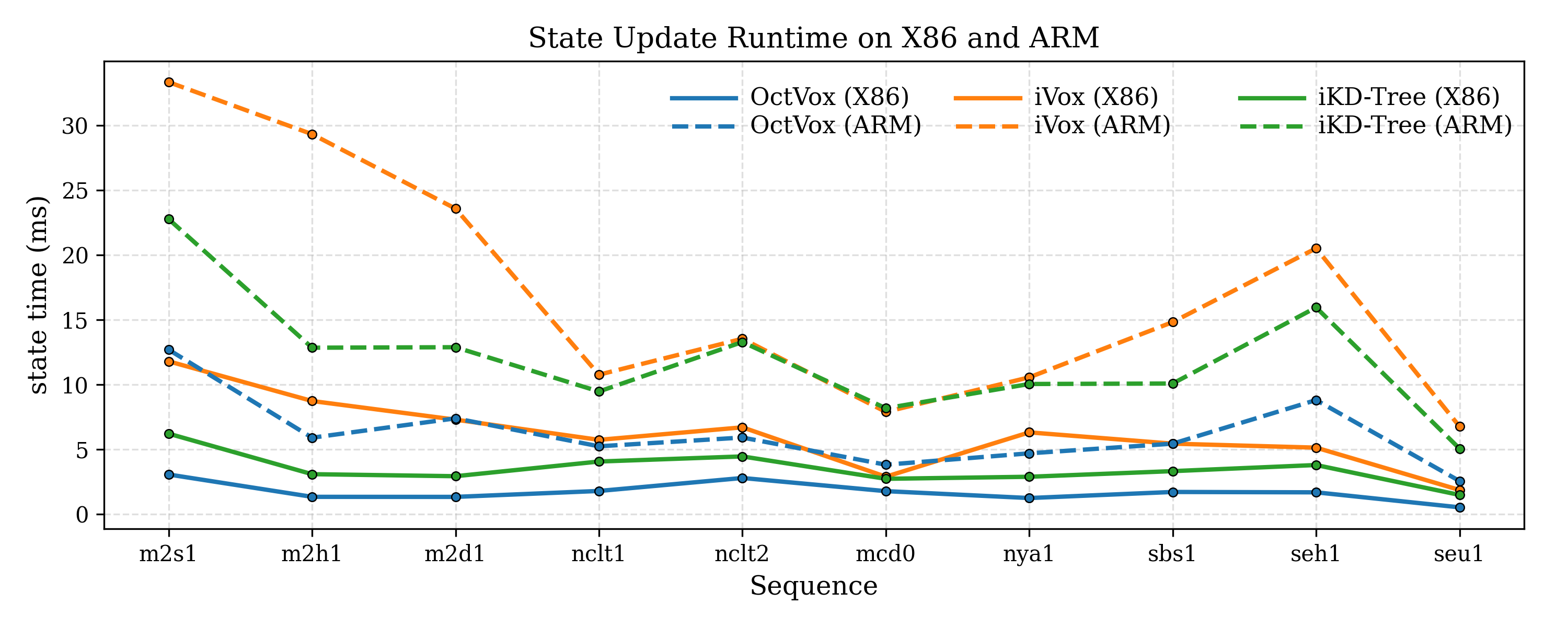}
\caption{State update time on representative sequences for OctVox, iVox~\cite{faster_lio}, and iKD-Tree~\cite{fast_lio2} on X86 and ARM.}
\label{fig:state_update}
\end{figure}

\begin{figure}[th]
\centering
\includegraphics[width=\columnwidth]{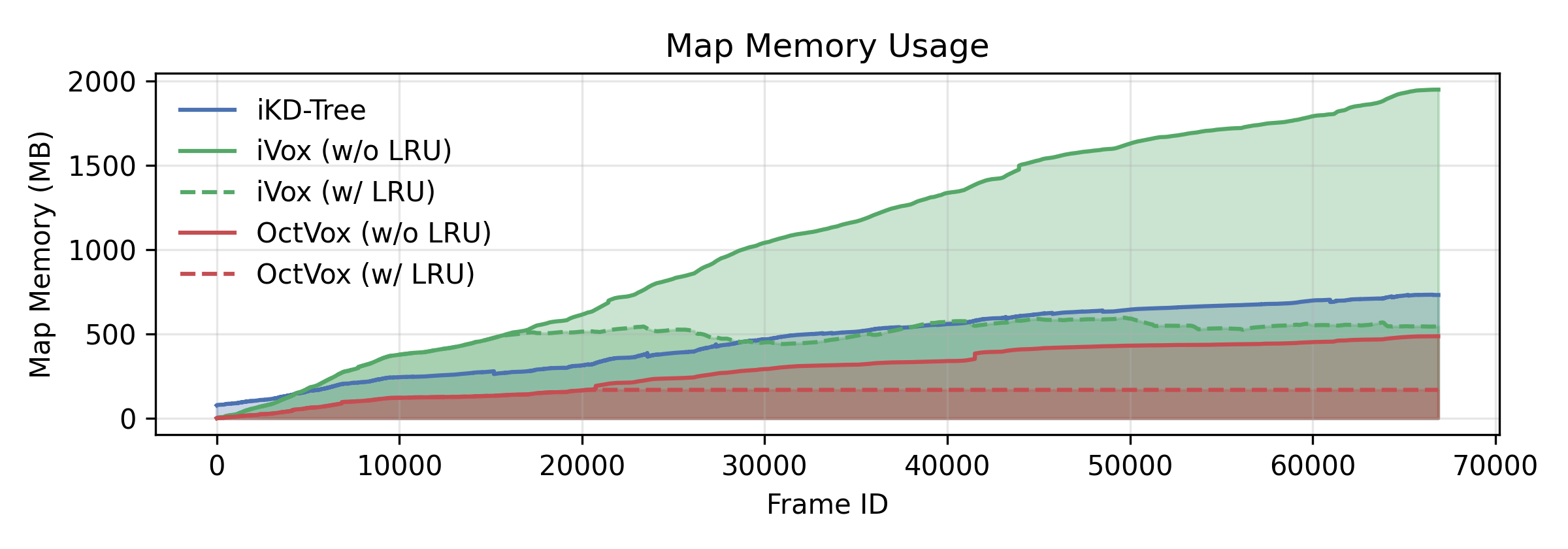}
\caption{Map memory usage along the frame index on the NCLT~1 sequence.}
\label{fig:map_memory_area}
\end{figure}

Figure~\ref{fig:module_time} shows the module-level timing across several representative scenes. 
Super-LIO maintains the best overall efficiency. The largest reduction comes from the \emph{state\_update} stage, highlighting the impact of OctVox and HKNN. 
Super-LIO also spends less time on \emph{undistortion} due to parallel processing, whereas iG-LIO incurs additional overhead in \emph{downsample} because of per-point covariance computations. 
Overall, the \emph{state\_update} stage dominates the total runtime across all frameworks. 

To isolate the effect of the proposed map representation, Figure~\ref{fig:state_update} replaces OctVox with iVox and iKD-Tree within the same Super-LIO pipeline and reports the average state update time on ten representative sequences for both X86 and ARM platforms. The improvements obtained with OctVox are clear under identical conditions. Finally, Figure~\ref{fig:map_memory_area} examines memory behavior on the longest sequence (nclt1, 1.8 hours). OctVox maintains the lowest and smoothest memory footprint without LRU, and with LRU enabled it stabilizes once reaching maximum capacity, while iVox shows minor memory variations.

\begin{figure}[]
\centering
\includegraphics[width=\columnwidth]{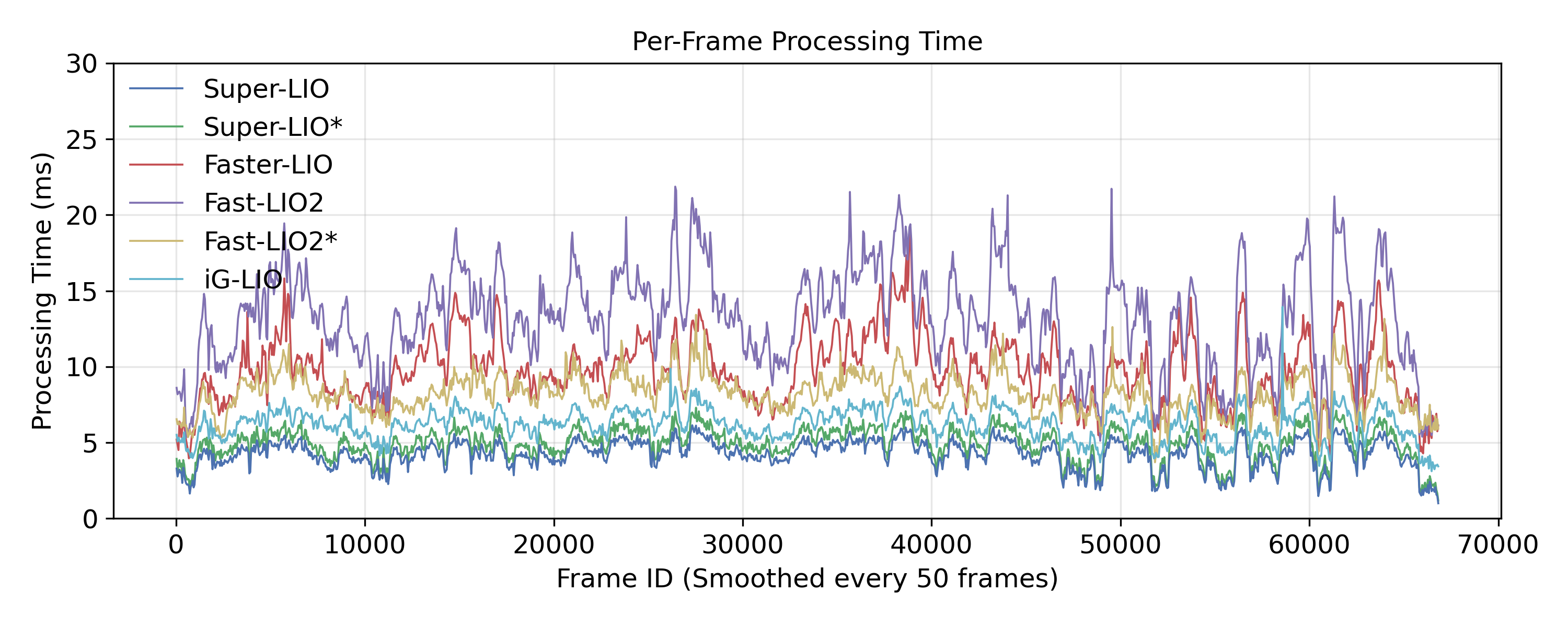}
\caption{Smoothed per-frame processing time on the NCLT~1 sequence.}
\label{fig:frame_time}
\end{figure}

\begin{figure}[]
\centering
\includegraphics[width=\columnwidth]{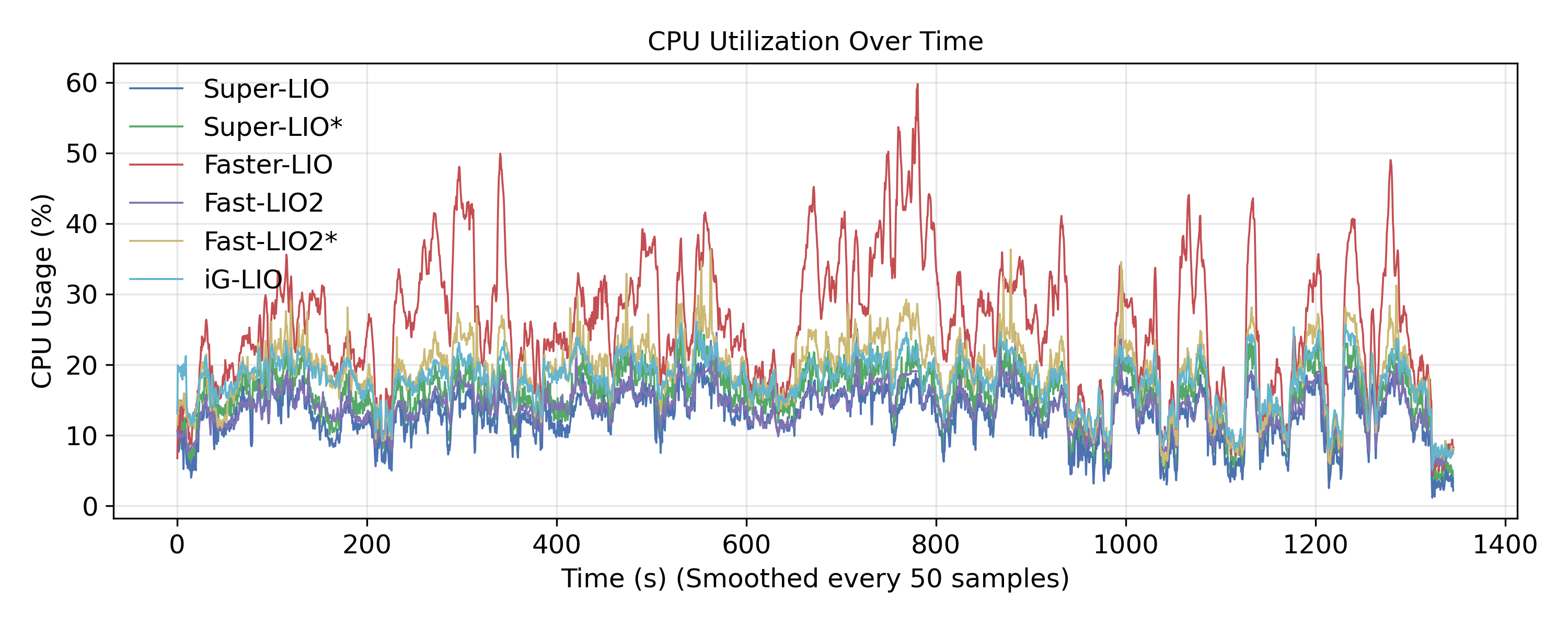}
\caption{CPU utilization over time on the NCLT~1 sequence.}
\label{fig:cpu_usage}
\end{figure}

Figures~\ref{fig:frame_time} and \ref{fig:cpu_usage} examine the per-frame processing time and CPU usage on the NCLT~1 sequence. Super-LIO shows a clearly lighter computational load across the run. These plots provide a clearer view of the real-time behavior of all methods.

\subsection{Robustness Analysis}
Our self-collected dataset covers diverse and challenging scenarios. In UAV sequences, the maximum linear velocity reaches 4.95\,m/s, the maximum angular velocity is 4.33\,rad/s, and the longest trajectory extends to 645\,m. Indoor runs include narrow corridors with widths below 1.5\,m, which are particularly demanding for real-time LiDAR-Inertial Odometry. Across all sequences, Super-LIO consistently maintains stable performance. 

Combined with the evaluations on public datasets, featuring high-speed motion, long-duration trajectories, diverse indoor and outdoor scenes, and sparse aerial point clouds, these results demonstrate that Super-LIO achieves robust and reliable performance under a wide spectrum of real-world operating conditions.

We also evaluated performance on both X86 and ARM platforms, confirming the cross-platform consistency of Super-LIO, as summarized in Tables~\ref{tab:x86_runtime} and \ref{tab:arm_runtime}. In contrast, iG-LIO shows competitive real-time performance on X86 but runs slower on ARM and fails on all NTU VIRAL sequences.

\section{Conclusion} \label{sec5}

We have introduced Super-LIO, a LiDAR-Inertial Odometry system that combines accuracy, efficiency, and robustness. Its design relies on two lightweight modules: OctVox for compact octo-voxel mapping and HKNN for fast and reliable KNN search. Extensive evaluations on public benchmarks and challenging self-collected datasets show consistently higher efficiency without compromising accuracy or robustness, including in high-speed, long-duration, sparse-sensing, and narrow-scene scenarios. Tests on both X86 and ARM platforms further validate stable performance and suitability for embedded deployment.

Super-LIO will be released as open source and is intended as a practical, integrable solution for robotic applications with strict compute and power limitations.

\bibliographystyle{IEEEtran}
\bibliography{bibliography}

\end{document}